\documentclass[a4paper, 11pt]{article}
\usepackage[T1]{fontenc}
\usepackage{authblk}
\usepackage{graphicx}
\graphicspath{{figures/}}
\usepackage{amsmath}
\usepackage{amsfonts}
\usepackage{graphicx}
\usepackage[colorlinks=true, allcolors=blue]{hyperref}
\usepackage{bm}
\usepackage[linesnumbered,ruled]{algorithm2e}
\usepackage{comment}
\usepackage{multicol}
\usepackage{paralist}
\usepackage{color}
\usepackage{subcaption}
\usepackage{fontawesome5}

\definecolor{dkgreen}{rgb}{0,0.6,0}

\begin{document}
\newcommand{\R}{\mathbb{R}}
\newcommand{\todo}[1]{\textcolor{red}{+ (TODO) #1 +}}
\newcommand{\del}[1]{\textcolor{yellow}{- (del) #1 -}}
\newcommand{\new}[1]{\textcolor{blue}{#1}}  
\newcommand{\container}[1]{\textit{C}$_{#1}$}
\newcommand{\pu}[1]{\textit{P}$_{#1}$}
\newcommand{\timeincrement}{\delta}
\newcommand{\rmin}{r_{\min}}
\newcommand{\rpen}{r_\text{pen}}
\newcommand{\benchmark}{ContainerGym}
\providecommand{\keywords}[1]{\small\textbf{\textit{Keywords---}} #1}
\title{\benchmark: A Real-World Reinforcement Learning Benchmark for Resource Allocation}
\author[1]{Abhijeet Pendyala}
\author[1]{Justin Dettmer}
\author[1]{Tobias Glasmachers}
\author[1]{Asma Atamna (\faIcon{envelope})}
\affil[1]{Ruhr-University Bochum, Bochum, Germany \protect\\
\tt firstname.lastname@ini.rub.de}
\setcounter{Maxaffil}{0}
\renewcommand\Affilfont{\itshape\small}
\date{}
\maketitle
\begin{abstract}
We present \benchmark, a benchmark for reinforcement learning inspired by a real-world industrial resource allocation task.
The proposed benchmark encodes a range of challenges commonly encountered in real-world sequential decision making problems, such as uncertainty. It can be configured to instantiate problems of varying degrees of difficulty, e.g., in terms of variable dimensionality.
Our benchmark differs from other reinforcement learning benchmarks, including the ones aiming to encode real-world difficulties, in that it is directly derived from a real-world industrial problem, which underwent minimal simplification and streamlining. It is sufficiently versatile to evaluate reinforcement learning algorithms on any real-world problem that fits our resource allocation framework.
We provide results of standard baseline methods. Going beyond the usual training reward curves, our results and the statistical tools used to interpret them allow to highlight interesting limitations of well-known deep reinforcement learning algorithms, namely PPO, TRPO and DQN.

\keywords{Deep reinforcement learning, Real-world benchmark, Resource allocation.}
\end{abstract}
\section{Introduction}
Supervised learning has long made its way into many industries, but industrial applications of (deep) reinforcement learning (RL) are significantly rare. This may be for many reasons, like the focus on impactful RL success stories in the area of games, a lower degree of technology readiness, and a lack of industrial RL benchmark problems.

A RL agent learns by taking sequential actions in its environment, observing the state of the environment, and receiving rewards~\cite{sutton2018reinforcement}. Reinforcement learning aims to fulfill the enticing promise of training a smart agent that solves a complex task through trial-and-error interactions with the environment, without specifying how the goal will be achieved. Great strides have been made in this direction, also in the real world, with notable applications in domains like robotics~\cite{DBLP:journals/corr/abs-1812-05905}, autonomous driving~\cite{9196730,DBLP:journals/corr/abs-1812-05905}, and control problems such as optimizing the power efficiency of data centers~\cite{47493}, control of nuclear fusion reactors~\cite{Degrave2022}, and optimizing gas turbines~\cite{Compare2018}. 

Yet, the accelerated progress in these areas has been fueled by making agents play in virtual gaming environments such as Atari 2600 games~\cite{dqn:mnih2013}, the game of GO~\cite{Silver2016}, and complex video games like Starcraft II~\cite{Vinyals2019}. These games provided sufficiently challenging environments to quickly test new algorithms and ideas, and gave rise to a suite of RL benchmark environments. The use of such environments to benchmark agents for industrial deployment comes with certain drawbacks. For instance, the environments either may not be challenging enough for the state-of-the-art algorithms (low dimensionality of state and action spaces) or require significant computational resources to solve. Industrial problems deviate widely from games in many further properties. Primarily, exploration in real-world systems often has strong safety constraints, and constitutes a balancing act between maximizing reward with good actions which are often sparse, and minimizing potentially severe consequences from bad actions. This is in contrast to training on a gaming environment, where the impact of a single action is often smaller, and the repercussions of poor decisions accumulate slowly over time.
In addition, underlying dynamics of gaming environments---several of which are near-deterministic---may not reflect the stochasticity of a real industrial system. Finally, the available environments may have a tedious setup procedure with restrictive licensing and dependencies on closed-source binaries.

To address these issues, we present {\benchmark}, an open-source real-world benchmark environment for RL algorithms. It is adapted from a digital twin of a high throughput processing industry. Our concrete use-case comes from a waste sorting application.
Our benchmark focuses on two phenomena of general interest: First, a stochastic model for a resource-filling process, where certain material is being accumulated in multiple storage containers. Second, a model for a resource transforming system, which takes in the material from these containers, and transforms it for further post-processing downstream. The processing units are a scarce resource, since they are large and expensive, and due to limited options of conveyor belt layout, only a small number can be used per plant.
{\benchmark} is not intended to be a perfect replica of a real system but serves the same hardness and complexity. The search for an optimal sequence of actions in {\benchmark} is akin to solving a dynamic resource allocation problem for the resource-transforming system, while also learning an optimal control behavior of the resource-filling process. In addition, the complexity of the environment is customizable. This allows testing the limitations of any given learning algorithm. This work aims to enable RL practitioners to quickly and reliably test their learning agents on an environment encoding real-world dynamics.

The paper is arranged as follows. Section~\ref{sec:related_work} discusses the relevant literature and motivates our contribution. Section~\ref{sec:rl_preliminaries} gives a brief introduction to reinforcement learning preliminaries. In Section~\ref{sec:container_management_env}, we present the real-world industrial control task that inspired {\benchmark} and discuss the challenges it presents. In Section~\ref{sec:formulation}, we formulate the real-world problem as a RL one and discuss the design choices that lead to our digital twin. We briefly present {\benchmark}'s implementation in Section~\ref{sec:software}. We present and discuss benchmark experiments of baseline methods in Section~\ref{sec:results}, and close with our conclusions in Section~\ref{sec:conclusion}.

\section{Related Work}\label{sec:related_work}
The majority of the existing open-source environments on which novel reinforcement learning algorithms could be tuned can be broadly divided into the following categories: toy control, robotics (MuJoCo)~\cite{6386109}, video games (Atari)~\cite{dqn:mnih2013}, and autonomous driving. The underlying dynamics of these environments are artificial and may not truly reflect real-world dynamic conditions like high dimensional states and action spaces, and stochastic dynamics. To the best of our knowledge, there exist very few such open-source benchmarks for industrial applications. To accelerate the deployment of agents in the industry, there is a need for a suite of RL benchmarks inspired by real-world industrial control problems, thereby making our benchmark environment, {\benchmark}, a valuable addition to it.

The classic control environments like mountain car, pendulum, or toy physics control environments based on Box2D are stochastic only in terms of their initial state. They have low dimensional state and action spaces, and they are considered easy to solve with standard methods. Also, the 50 commonly used Atari games in the Arcade Learning Environment~\cite{Bellemare2013}, where nonlinear control policies need to be learned, are routinely solved to a super-human level by well-established algorithms. This environment, although posing high dimensionality, is deterministic. The real world is not deterministic and there is a need to tune algorithms that can cope with stochasticity. Although techniques like sticky actions or skipping a random number of initial frames have been developed to add artificial randomness, this randomness may still be very structured and not challenging enough. On the other hand, video game simulators like Starcraft II~\cite{Vinyals2019} offer high-dimensional image observations, partial observability, and (slight) stochasticity. However, playing around with DeepRL agents on such environments requires substantial computational resources. It might very well be overkill to tune reinforcement learning agents in these environments when the real goal is to excel in industrial applications.

Advancements in the RL world, in games like Go and Chess, were achieved by exploiting the rules of these games and embedding them into a stochastic planner. In real-world environments, this is seldom achievable, as these systems are highly complex to model in their entirety. Such systems are modeled as partially observable Markov decision processes and present a tough challenge for learning agents that can explore only through interactions. Lastly, some of the more sophisticated RL environments available, e.g., advanced physics simulators like MuJoCo~\cite{6386109}, offer licenses with restrictive terms of use. Also, environments like Starcraft II require access to a closed-source binary. Open source licensing in an environment is highly desirable for RL practitioners as it enables them to debug the code, extend the functionality and test new research ideas.

Other related works, amongst the very few available open-source reinforcement learning environments for industrial problems, are Real-world RL suite~\cite{dulacarnold2020realworldrlempirical} and Industrial benchmark (IB)~\cite{Hein_2017} environments. Real-world RL-suite is not derived from a real-world scenario, rather the existing toy problems are perturbed to mimic the conditions in a real-world problem. The IB comes close in spirit to our work, although it lacks the customizability of {\benchmark} and our expanded tools for agent behavior explainability. Additionally, the (continuous) action and state spaces are relatively low dimensional and of fixed sizes. 

\section{Reinforcement Learning Preliminaries}\label{sec:rl_preliminaries}
RL problems are typically studied as discrete-time Markov decision processes (MDPs), where a MDP can be defined as a tuple $\langle \mathcal{S}, \mathcal{A}, p, r, \gamma \rangle$. At timestep $t$, the agent is in a state $s_t \in \mathcal{S}$ and takes an action $a_t \in \mathcal{A}$. It arrives in a new state $s_{t + 1}$ with probability $p(s_{t + 1} \mid s_t, a_t)$ and receives a reward $r(s_t, a_t, s_{t + 1})$. The state transitions of a MDP satisfy the Markov property $p(s_{t + 1} \mid s_t, a_t, \dots, s_0, a_0) = p(s_{t + 1} \mid s_t, a_t)$. That is, the new state $s_{t + 1}$ only depends on the current state $s_t$ and action $a_t$.
The goal of the RL agent interacting with the MDP is to find a policy $\pi: \mathcal{S} \to \mathcal{A}$ that maximizes the expected (discounted) cumulative reward. This optimization problem is defined formally as follows:
\begin{equation}\label{eq:optimal_policy}
    \arg \max_\pi \mathbb{E}_{\tau \sim \pi} \left[\sum_{t \geq 0} \gamma^t r(s_t, a_t, s_{t + 1})\right] \enspace ,
\end{equation}
where $\tau = (s_0, a_0, r_0, s_1, \dots)$ is a trajectory generated by following $\pi$ and $\gamma \in (0, 1]$ is a discount factor. A trajectory $\tau$ ends either when a maximum timestep count $T$---also called episode\footnote{We use the terms ``episode'' and ``rollout'' interchangeably in this paper.} length---is reached, or when a terminal state is reached (early termination).

\section{Container Management Environment}\label{sec:container_management_env}
In this section, we describe the real-world industrial control task that inspired our RL benchmark. It originates from the final stage of a waste sorting process.

The environment consists of a solid material-transforming facility that hosts a set of \textit{containers} and a significantly smaller set of \textit{processing units} (PUs).
Containers are continuously filled with material, where the material flow rate is a container-dependent stochastic process. They must be emptied regularly so that their content can be transformed by the PUs.
When a container is emptied, its content is transported on a conveyor belt to a free PU that transforms it into \textit{products}. It is not possible to extract material from a container without emptying it completely.
The number of produced products depends on the volume of the material being processed. Ideally, this volume should be an integer multiple of the product's size. Otherwise, the surplus volume that cannot be transformed into a product is redirected to the corresponding container again via an energetically costly process that we do not consider in this work. 
Each container has at least one optimal emptying volume: a global optimum and possibly other, smaller, local optima that are all multiples of container-specific product size. Generally speaking, larger volumes are better, since PUs are more efficient with producing many products in a series.

The worst-case scenario is a container overflow. In the waste sorting application inspiring our benchmark, it incurs a high recovery cost including human intervention to stop and restart the facility. This situation is undesirable and should be actively avoided. Therefore, letting containers come close to their capacity limit is rather risky.

The quality of an emptying decision is a compromise between the number of potential products and the costs resulting from actuating a PU and handling surplus volume. Therefore, the closer an emptying volume is to an optimum, the better. If a container is emptied too far away from any optimal volume, the costs outweigh the benefit of producing products.

This setup can be framed more broadly as a resource allocation problem, where one item of a scarce resource, namely the PUs, needs to be allocated whenever an emptying decision is made. If no PU is available, the container cannot be emptied and will continue to fill up.

\noindent There are multiple aspects that make this problem challenging:
\begin{compactitem}
    \item The rate at which the material arrives at the containers is stochastic. Indeed, although the volumes follow a globally linear trend---as discussed in details in Section~\ref{sec:formulation}, the measurements can be very noisy. The influx of material is variable, and there is added noise from the sensor readings inside the containers. This makes applying standard planning approaches difficult.
    \item The scarcity of the PUs implies that always waiting for containers to fill up to their ideal emptying volume is risky: if no PU is available at that time, then we risk an overflow. This challenge becomes more prominent when the number of containers---in particular, the ratio between the number of containers and the number of PUs---increases. Therefore, an optimal policy needs to take fill states and fill rates of all containers into account, and possibly empty some containers early.
    \item Emptying decisions can be taken at any time, but in a close-to-optimal policy, they are rather infrequent. Also, the rate at which containers should be emptied varies between containers. Therefore, the distributions of actions are highly asymmetric, with important actions (and corresponding rewards) being rare.
\end{compactitem}

\section{Reinforcement Learning Problem Formulation}\label{sec:formulation}
We formulate the container management problem presented in Section~\ref{sec:container_management_env} as a MDP that can be addressed by RL approaches. The resulting MDP is an accurate representation of the original problem, as only mild simplifications are made. Specifically, we model one category of PUs instead of two, we do not include inactivity periods of the facility in our environment, and we neglect the durations of processes of minor relevance. All parameters described in the rest of this section are estimated from real-world data (system identification). Overall, the MDP reflects the challenges properly without complicating the benchmark (and the code base) with irrelevant details.

\subsection{State and Action Spaces}
\subsubsection{State space}
The state $s_t$ of the system at any given time $t$ is defined by the volumes $v_{i, t}$ of material contained in each container \container{i}, and a timer $p_{j, t}$ for each PU \pu{j} indicating in how many seconds the PU will be ready to use. A value of zero means that the PU is available, while a positive value means that it is currently busy. Therefore, $s_t = (\{v_{i, t}\}_{i = 1}^n, \{p_{j, t}\}_{j = 1}^m)$, where $n$ and $m$ are the number of containers and PUs, respectively. Valid initial states include non-negative volumes not greater than the maximum container capacity $v_{\max}$, and non-negative timer values.

\subsubsection{Action space}
Possible actions at a given time $t$ are either (i) not to empty any container, i.e. to do nothing, or (ii) to empty a container \container{i} and transform its content using one of the PUs.
The action of doing nothing is encoded with $0$, whereas the action of emptying a container \container{i} and transforming its content is encoded with the container's index $i$. Therefore, $a_t \in \{0, 1, \dots, n\}$, where $n$ is the number of containers.
Since we consider identical PUs, specifying which PU is used in the action encoding is not necessary as an emptying action will result in the same state for all the PUs, given they were at the same previous state.

\subsection{Environment Dynamics}
In this section, we define the dynamics of the volume of material in the containers, the PU model, as well as the state update.

\subsubsection{Volume dynamics}
The volume of material in each container increases following an irregular trend, growing linearly on average, captured by a random walk model with drift. That is, given the current volume $v_{i, t}$ contained in \container{i}, the volume at time $t + 1$ is given by the function $f_i$ defined as
\begin{equation}\label{eq:random_walk}
    f_i(v_{i, t}) = \max(0, \alpha_i + v_{i, t} + \epsilon_{i, t}) \enspace ,
\end{equation}
where $\alpha_i > 0$ is the slope of the linear upward trend followed by the volume for \container{i} and the noise $\epsilon_{i, t}$ is sampled from a normal distribution $\mathcal{N}(0, \sigma_i^2)$ with mean $0$ and variance $\sigma_i^2$. The $\max$ operator forces the volume to non-negative values.

When a container \container{i} is emptied, its volume drops to $0$ at the next timestep. Although the volume drops progressively to $0$ in the real facility, empirical evidence provided by our data shows that emptying durations are within the range of the time-step lengths considered in this paper ($60$ and $120$ seconds).

\subsubsection{Processing unit dynamics}
The time (in seconds) needed by a PU to transform a volume $v$ of material in a container \container{i} is linear in the number of products $\lfloor v / b_i \rfloor$ that can be produced from $v$. It is given by the function $g_i$ defined in Equation~\eqref{eq:pu_model}, where $b_i > 0$ is the product size, $\beta_i > 0$ is the time it takes to actuate the PU before products can be produced,
and $\lambda_i > 0$ is the time per product. Note that all the parameters indexed with $i$ are container-dependent.
\begin{equation}\label{eq:pu_model}
    g_i(v) = \beta_i + \lambda_i \lfloor v / b_i \rfloor \enspace .
\end{equation}
A PU can only produce one type of product at a time. Therefore, it can be used for emptying a container only when it is free. Therefore, if all PUs are busy, the container trying to use one is not emptied and continues to fill up.

\subsubsection{State update}
We distinguish between the following cases to define the new state $s_{t + 1} = (\{v_{i, t + 1}\}_{i = 1}^n, \{p_{j, t + 1}\}_{j = 1}^m)$ given the current state $s_t$ and action $a_t$.

\paragraph{$\bm{a_t = 0}$.}
This corresponds to the action of doing nothing. The material volumes inside the containers increase while the timers indicating the availability of the PUs are decreased according to:
\begin{align}
    v_{i, t + 1} &= f_i(v_{i, t}), & i \in \{1, \dots, n\} \enspace , \label{eq:vol_update_a_0} \\
    p_{j, t + 1} &= \max(0, p_{j, t} - \timeincrement), & j \in \{1, \dots, m\} \enspace , \label{eq:pu_update_a_0}
\end{align}
where $f_i$ is the random walk model defined in Equation~\eqref{eq:random_walk} and $\timeincrement$ is the length of a timestep in seconds.

\paragraph{$\bm{a_t \neq 0}$.}
This corresponds to an emptying action. If no PU is available, that is, $p_{j, t} > 0 \enspace \forall \, j = 1, \dots, m$, the updates are identical to the one defined in Equations~\eqref{eq:vol_update_a_0} and~\eqref{eq:pu_update_a_0}. If at least one PU \pu{k} is available, the new state variables are defined as follows:
\begin{align}
    v_{a_t, t + 1} &= 0 \enspace , \\
    v_{i, t + 1} &= f_i(v_{i, t}), & i \in \{1, \dots, n\} \backslash \{a_t\} \enspace , \\
    p_{k, t + 1} &= g_{a_t} (v_{a_t, t}) \enspace , \\
    p_{j, t + 1} &= \max(0, p_{j, t} - \timeincrement), & j \in \{1, \dots, m\} \backslash \{k\} \enspace ,
\end{align}
where the value of the action $a_t$ is the index of the container \container{a_t} to empty, `$\backslash$' denotes the set difference operator, $f_i$ is the random walk model defined in Equation~\eqref{eq:random_walk}, and $g_{a_t}$ is the processing time defined in Equation~\eqref{eq:pu_model}.

Although the processes can continue indefinitely, we stop an episode after a maximum length of $T$ timesteps. This is done to make {\benchmark} compatible with RL algorithms designed for episodic tasks, and hence to maximize its utility. When a container reaches its maximum volume $v_{\max}$, however, the episode is terminated and a negative reward is returned (see details in Section~\ref{subsec:reward_fct}).

\subsection{Reward Function}\label{subsec:reward_fct}
We use a deterministic reward function $r(s_t, a_t, s_{t + 1})$ where higher values correspond to better $(s_t, a_t)$ pairs. The new state $s_{t + 1}$ is taken into account to return a large negative reward $r_\text{min}$ when a container overflows, i.e. $\exists i \in \{1, \dots, n\}$, $v_{i, t + 1} \geq v_{\max}$, before ending the episode. In all other cases, the immediate reward is determined only by the current state $s_t$ and the action~$a_t$.

\paragraph{$\bm{a_t = 0}$.}
If the action is to do nothing, we define $r(s_t, 0, s_{t + 1}) = 0$.

\begin{figure}
\centering
\includegraphics[width=0.45\textwidth]{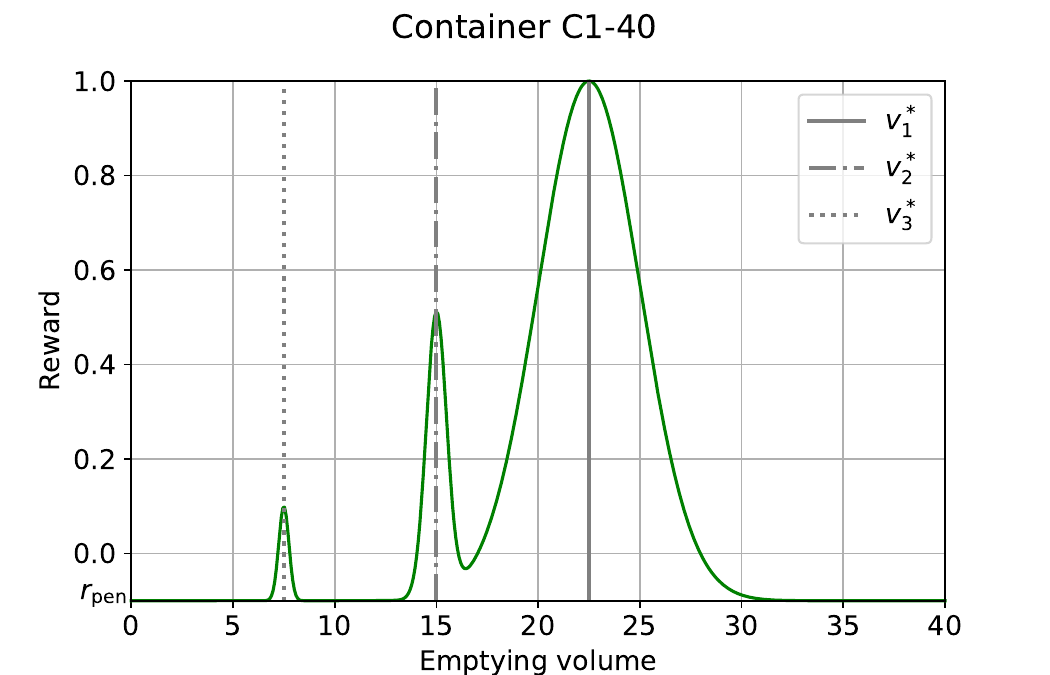}
\caption{Rewards received when emptying a container with three optima. The design of the reward function fosters emptying late, hence allowing PUs to produce many products in a row.} \label{fig:gaussian_emptying_reward}
\end{figure}

\paragraph{$\bm{a_t \neq 0}$.}
If an emptying action is selected while (i) no PU is available or (ii) the selected container is already empty, i.e. $v_{a_t, t} = 0$, then $r(s_t, a_t, s_{t + 1}) = \rpen$, where it holds $\rmin < \rpen < 0$ for the penalty. If, on the other hand, at least one PU is available and the volume to process is non-zero ($v_{a_t, t} > 0$), the reward is a finite sum of Gaussian functions centered around optimal emptying volumes $v_{a_t, i}^*, i = 1, \dots, p_{a_t}$, where the height of a peak $0 < h_{a_t, i} \leq 1$ is proportional to the quality of the corresponding optimum. The reward function in this case is defined as follows:
\begin{equation}\label{eq:gaussian_reward}
    r(s_t, a_t, s_{t + 1}) = r_\text{pen} + \sum_{i = 1}^{p_{a_t}} (h_{a_t, i} - r_\text{pen}) \exp\left(- \frac{(v_{a_t, t} - v_{a_t, i}^*)^2}{2 w_{a_t, i}^2}\right) \enspace ,
\end{equation}
where $p_{a_t}$ is the number of optima for container \container{a_t} and $w_{a_t, i} > 0$ is the width of the bell around $v_{a_t, i}^*$.
The Gaussian reward defined in Equation~\eqref{eq:gaussian_reward} takes its values in $\left]\rpen, 1\right]$, the maximum value $1$ being achieved at the ideal emptying volume $v_{a_t, 1}^*$ for which $h_{a_t, 1} = 1$.
The coefficients $h_{a, i}$ are designed so that processing large volumes at a time is beneficial, hence encoding a conflict between emptying containers early and risking overflow. This tension, together with the limited availability of the PUs, yields a highly non-trivial control task. Figure~\ref{fig:gaussian_emptying_reward} shows an example of Gaussian rewards when emptying a container with three optimal volumes at different volumes in $[0, 40[$.

\section{{\benchmark} Usage Guide}\label{sec:software}
In this section, we introduce the OpenAI Gym implementation\footnote{The \benchmark{} software is available on the following GitHub repository: \url{https://github.com/Pendu/ContainerGym}.} of our benchmark environment. We present the customizable parameters of the benchmark, provide an outline for how the Python implementation reflects the theoretical definition in Section~\ref{sec:formulation}, and show which tools we provide to extend the understanding of an agent's behavior beyond the achieved rewards.

\paragraph{Gym implementation.}
Our environment follows the OpenAI Gym~\cite{gym1606.01540} framework.
It implements a \texttt{step()}-function computing a state update.
The \texttt{reset()}-function resets time to $t=0$ and returns the environment to a valid initial state, and the \texttt{render()}-function displays a live diagram of the environment.

We deliver functionality to configure the environment's complexity and difficulty through its parameters such as the number of containers and PUs, as well as the composition of the reward function, the overflow penalty $\rmin$, the sub-optimal emptying penalty $\rpen$, the length of a timestep $\delta$, and the length of an episode~$T$.
We provide example configurations in our GitHub repository that are close in nature to the real industrial facility.

In the spirit of open and reproducible research, we include scripts and model files for reproducing the experiments presented in the next section.%

\paragraph{Action explainability.}
While most RL algorithms treat the environment as a black box, facility operators want to ``understand'' a policy before deploying it for production. To this end, the accumulated reward does not provide sufficient information, since it treats all mistakes uniformly. Practitioners need to understand which types of mistakes a sub-optimal policy makes. For example, a low reward can be obtained by systematically emptying containers too early (local optimum) or by emptying at non-integer multiples of the product size. When a basic emptying strategy for each container is in place, low rewards typically result from too many containers reaching their ideal volume at the same time so that PUs are overloaded. To make the different types of issues of a policy transparent, we plot all individual container volumes over an entire episode. We further provide tools to create empirical cumulative distribution function plots over the volumes at which containers were emptied over multiple episodes. The latter plots in particular provide insights into the target volumes an agent aims to hit, and whether it does so reliably.

\section{Performance of Baseline Methods}\label{sec:results}
We illustrate the use of \benchmark{} by benchmarking three popular deep RL algorithms: two on-policy methods, namely Proximal Policy Optimization (PPO)~\cite{ppo:schulman2017} and Trust Region Policy Optimization (TRPO)~\cite{trpo:schulman2015}, and one off-policy method, namely Deep Q-Network (DQN)~\cite{dqn:mnih2013}. We also compare the performance of these RL approaches against a naive rule-based controller. By doing so, we establish an initial baseline on \benchmark. We use the Stable Baselines3 implementation of PPO, TRPO, and DQN~\cite{stable-baselines3}.

\subsection{Experimental Setup}
The algorithms are trained on $8$ \benchmark{} instances, detailed in Table~\ref{tab:test_cumul_rewards}, where the varied parameters are the number of containers $n$, the number of PUs $m$ and the timestep length $\timeincrement$.\footnote{Increasing the timestep length $\timeincrement$ should be done carefully. Otherwise, the problem could become trivial. In our case, we choose $\timeincrement$ such that it is smaller than the minimum time it takes a PU to process the volume equivalent to one product.}
The rationale behind the chosen configurations is to assess (i) how the algorithms scale with the environment dimensionality, (ii) how action frequency affects the trained policies and (iii) whether the optimal policy can be found in the conceptually trivial case $m = n$, where there is always a free PU when needed.
This is only a control experiment for testing RL performance. In practice, PUs are a scarce resource ($m \ll n$).

The maximum episode length $T$ is set to $1500$ timesteps during training in all experiments, whereas the initial volumes $v_{i, 0}$ are uniformly sampled in $[0, 30]$, and the maximum capacity is set to $v_{\max} = 40$ volume units for all containers. The initial PU states are set to free ($p_{j, 0} = 0$) and the minimum and penalty rewards are set to $\rmin = -1$ and $\rpen = -0.1$ respectively.
The algorithms are trained with an equal budget of $2$ (resp. $5$) million timesteps when $n = 5$ (resp. $n = 11$) and the number of steps used for each policy update for PPO and TRPO is $6144$. Default values are used for the remaining hyperparameters, as the aim of our work is to show characteristics of the training algorithms with a reasonable set of parameters. For each algorithm, $15$ policies are trained in parallel, each with a different seed $\in \{1, \dots, 15\}$, and the policy with the best training cumulative reward is returned for each run.
To make comparison easier, policies are evaluated on a similar test environment with $\timeincrement = 120$ and $T = 600$.

\subsection{Results and Discussion}
Table~\ref{tab:test_cumul_rewards} shows the average test cumulative reward, along with its standard deviation, for the best and median policies trained with PPO, TRPO, and DQN on each environment configuration. These statistics are calculated by running each policy $15$ times on the corresponding test environment.

\begin{table}[ht]
    \caption{Test cumulative reward, averaged over $15$ episodes, and its standard deviation for the best and median policies for PPO, TRPO and DQN. Best and median policies are selected from a sample of $15$ policies (seeds) trained on the investigated environment configurations. The highest best performance is highlighted for each configuration.}
    \centering
    \resizebox{\columnwidth}{!}{
    \begin{tabular}{c|c|c|c|c|c|c|c|c}
        \multicolumn{3}{c|}{Config.} & \multicolumn{2}{c|}{PPO} & \multicolumn{2}{c|}{TRPO} & \multicolumn{2}{c}{DQN} \\
        \hline
        $n$ & $m$ & $\timeincrement$ & best & median & best & median & best & median \\
        \hline
        5 & 2 & 60 & $\bm{38.55 \pm 1.87}$ & $29.94 \pm 8.42$ & $37.35 \pm 7.13$ & $4.93 \pm 4.90$ & $29.50 \pm 3.43$ & $1.57 \pm 0.93$ \\
        5 & 2 & 120 & $\bm{51.43 \pm 3.00}$ & $49.81 \pm 1.56$ & $38.33 \pm 8.36$ & $16.86 \pm 2.78$ & $42.96 \pm 2.80$ & $7.90 \pm 4.72$ \\
        5 & 5 & 60 & $\bm{37.54 \pm 1.50}$ & $30.64 \pm 2.44$ & $33.62 \pm 7.76$ & $7.36 \pm 4.40$ & $23.98 \pm 13.17$ & $1.96 \pm 1.28$ \\
        5 & 5 & 120 & $\bm{50.42 \pm 2.57}$ & $47.23 \pm 1.89$ & $47.36 \pm 2.07$ & $5.39 \pm 4.38$ & $43.26 \pm 3.73$ & $8.56 \pm 4.12$ \\
        11 & 2 & 60 & $31.01 \pm 8.62$ & $23.25 \pm 16.79$ & $26.62 \pm 18.94$ & $4.26 \pm 3.81$ & $\bm{42.63 \pm 21.73}$ & $11.84 \pm 10.33$ \\
        11 & 2 & 120 & $54.30 \pm 8.48$ & $49.58 \pm 13.54$ & $45.78 \pm 18.42$ & $32.08 \pm 18.04$ & $\bm{72.19 \pm 16.20}$ & $24.07 \pm 13.68$ \\
        11 & 11 & 60 & $27.87 \pm 8.89$ & $17.57 \pm 13.99$ & $15.66 \pm 10.65$ & $4.90 \pm 2.68$ & $\bm{28.32 \pm 22.42}$ & $8.49 \pm 4.92$ \\
        11 & 11 & 120 & $47.75 \pm 10.78$ & $42.37 \pm 13.25$ & $\bm{50.55 \pm 11.08}$ & $34.29 \pm 16.00$ & $29.06 \pm 13.10$ & $13.16 \pm 8.32$ \\
    \end{tabular}
    }
    \label{tab:test_cumul_rewards}
\end{table}

PPO achieves the highest cumulative reward on environments with $5$ containers. DQN, on the other hand, shows the best performance on higher dimensionality environments ($11$ containers), with the exception of the last configuration. It has, however, a particularly high variance. Its median performance is also significantly lower than the best one, suggesting less stability than PPO. DQN's particularly high rewards when $n = 11$ could be due to a better exploration of the search space. An exception is the last configuration, where DQN's performance is significantly below those of TRPO and PPO. Overall, PPO has smaller standard deviations and a smaller difference between best and median performances, suggesting higher stability than TRPO and DQN.

\begin{figure}
\centering
\includegraphics[clip, trim=3.3cm 0.6cm 0.3cm 0.3cm, width=0.49\textwidth]{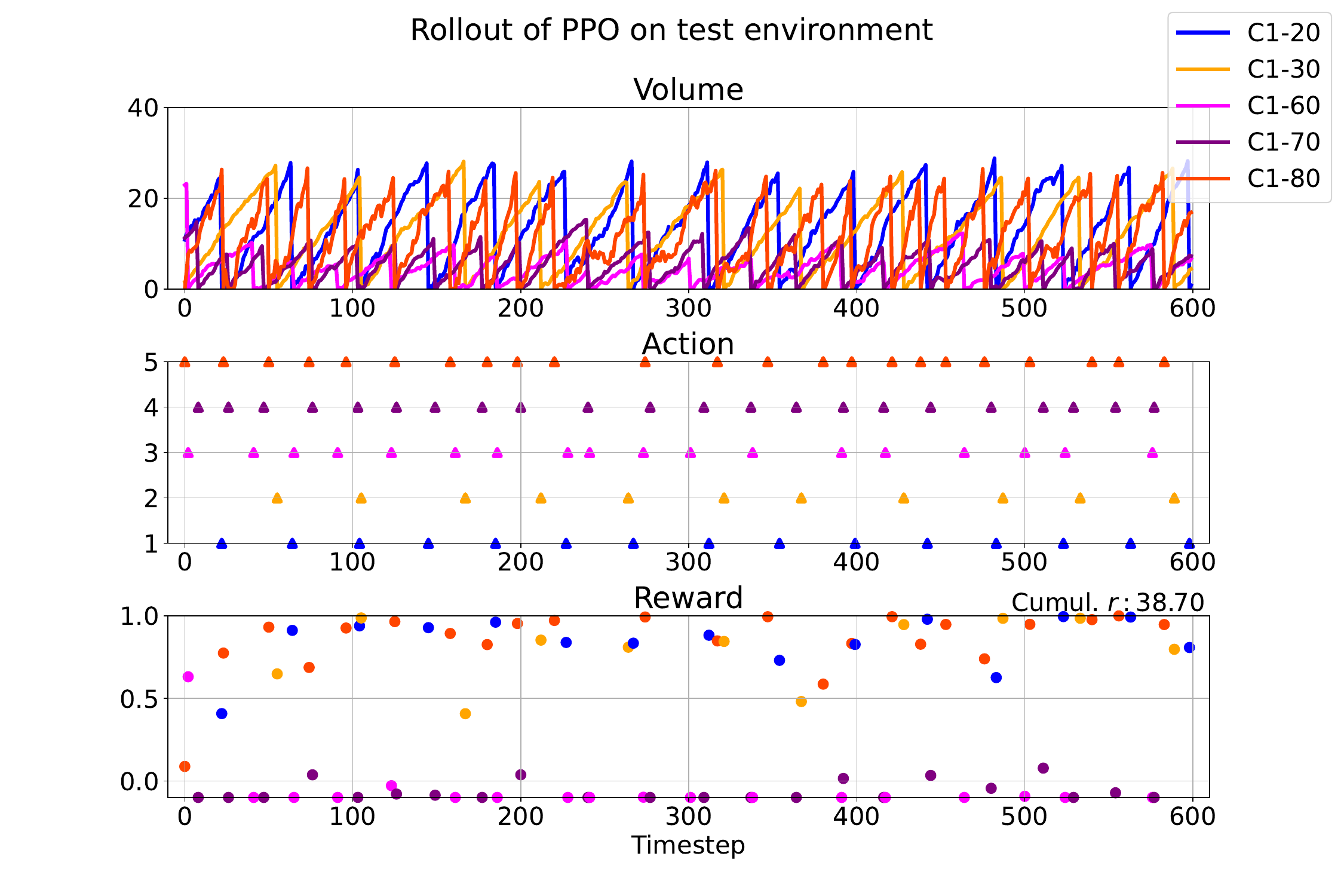}
\includegraphics[clip, trim=3.3cm 0.6cm 0.3cm 0.3cm, width=0.49\textwidth]{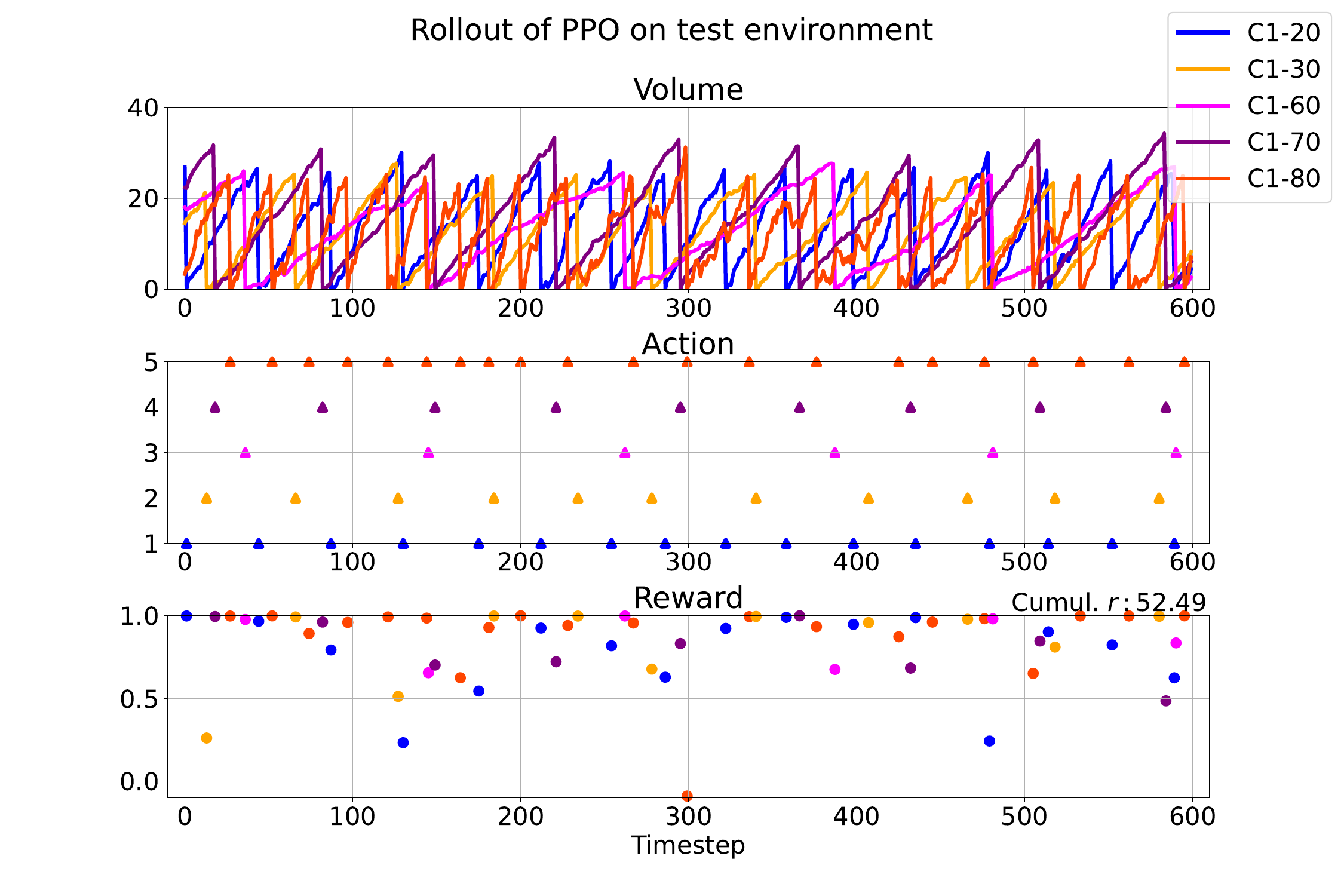} \\
\includegraphics[clip, trim=3.3cm 0.6cm 0.3cm 0.3cm, width=0.49\textwidth]{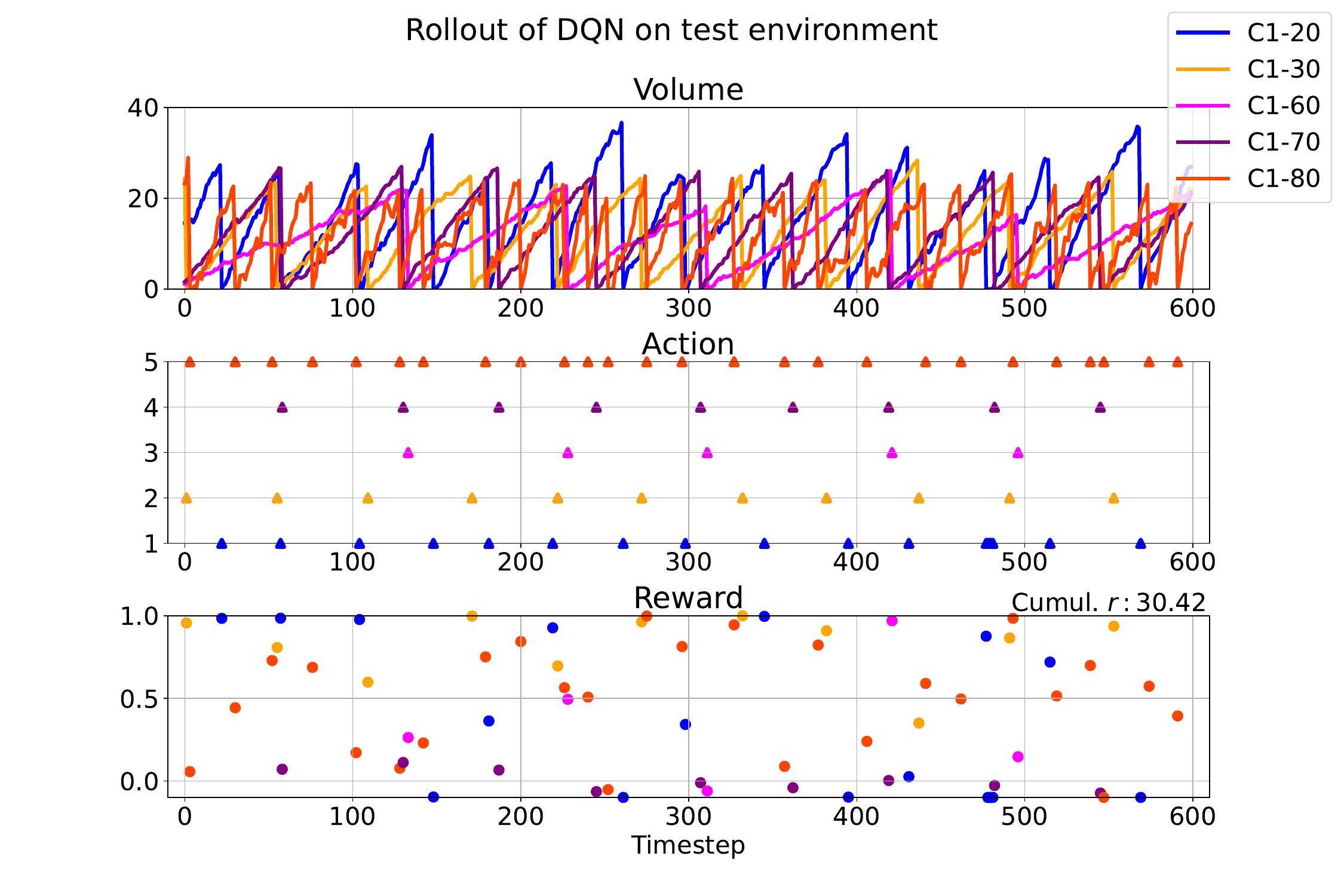}
\includegraphics[clip, trim=3.3cm 0.6cm 0.3cm 0.3cm, width=0.49\textwidth]{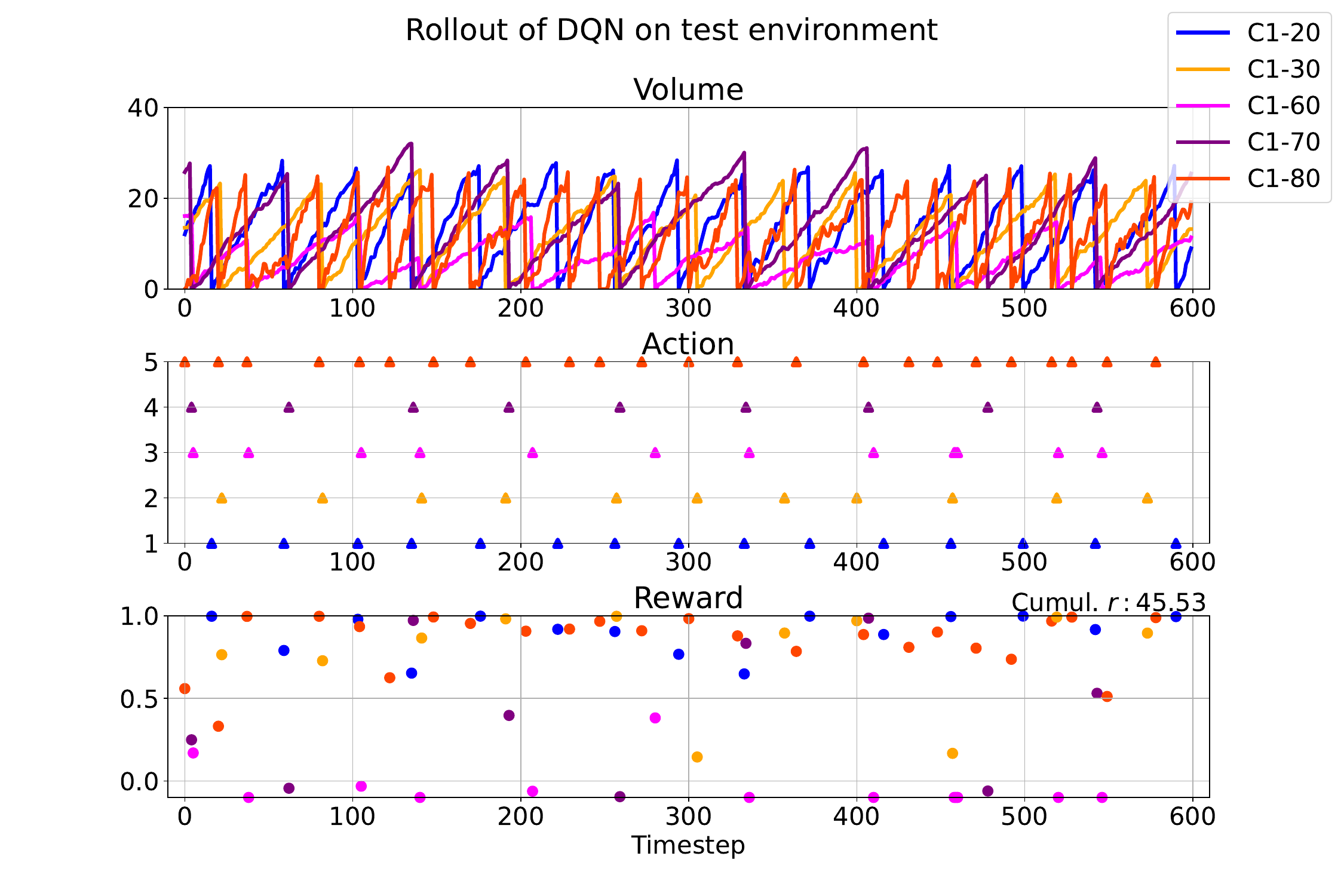}
\caption{Rollouts of the best policy trained with PPO (first row) and DQN (second row) on a test environment with $n = 5$ and $m = 2$. Left: training environment with $\timeincrement = 60$ seconds. Right: training environment with $\timeincrement = 120$. Displayed are the volumes, emptying actions and non-zero rewards at each timestep.} \label{fig:test_curves_n5_m2}
\end{figure}

Our results also show that taking actions at a lower frequency ($\timeincrement = 120$) leads to better policies. Due to space limitations, we focus on analyzing this effect on configurations with $n = 5$ and $m = 2$. Figure~\ref{fig:test_curves_n5_m2} displays the volumes, actions, and rewards over one episode for PPO and DQN when the (best) policy is trained with $\timeincrement = 60$ (left column) and with $\timeincrement = 120$.

\begin{figure}
\centering
\includegraphics[clip, trim=1cm 0.5cm 1cm 0.5cm, width=0.24\textwidth]{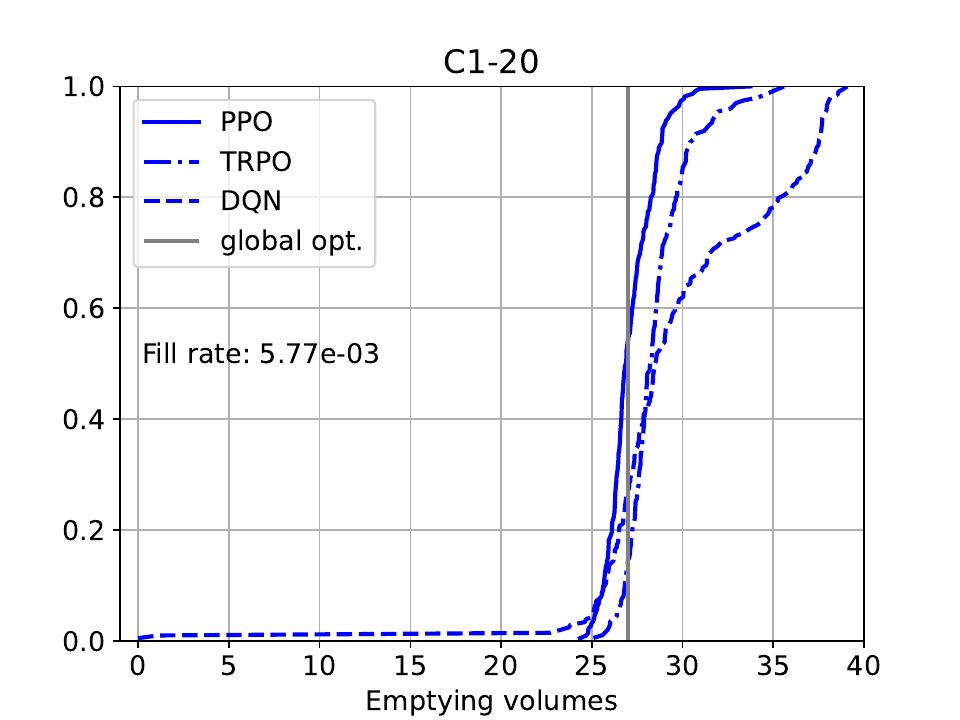}
\includegraphics[clip, trim=1cm 0.5cm 1cm 0.5cm, width=0.24\textwidth]{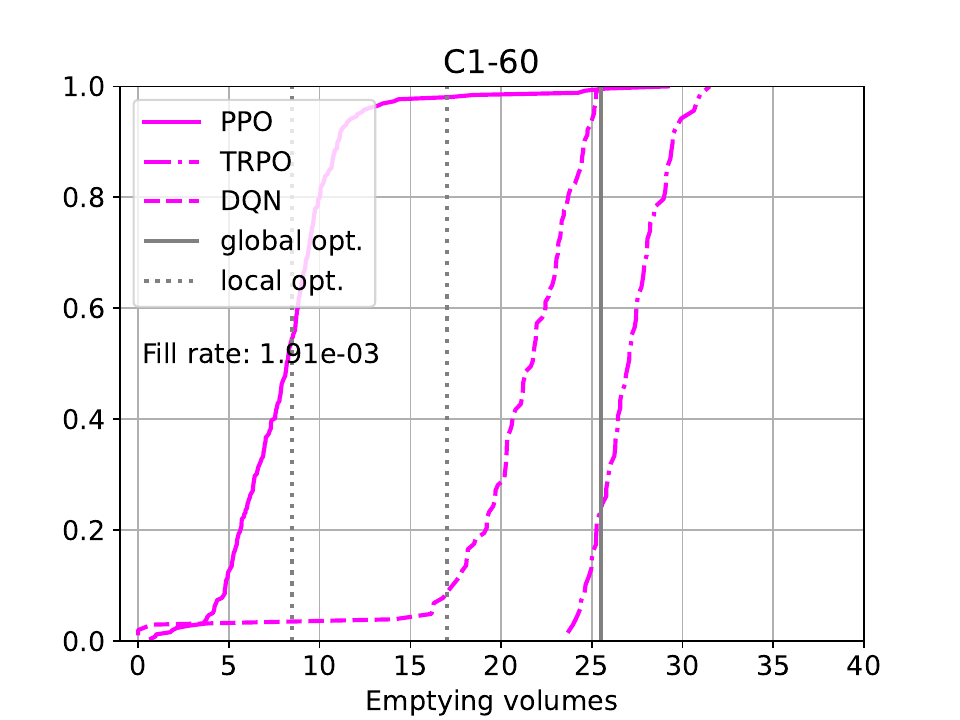}
\includegraphics[clip, trim=1cm 0.5cm 1cm 0.5cm, width=0.24\textwidth]{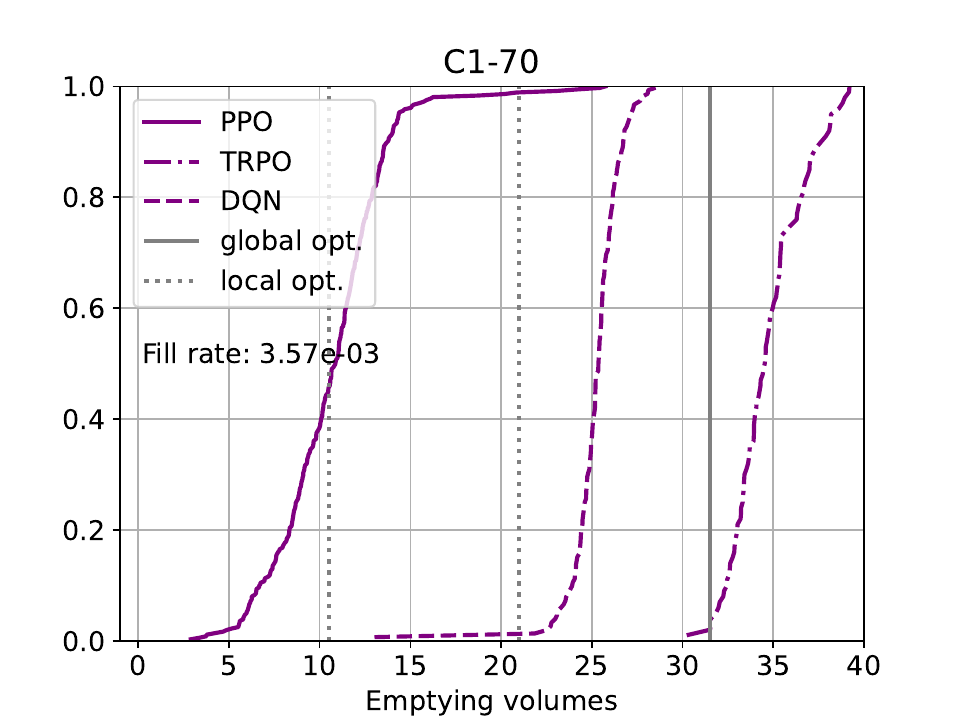}
\includegraphics[clip, trim=1cm 0.5cm 1cm 0.5cm, width=0.24\textwidth]{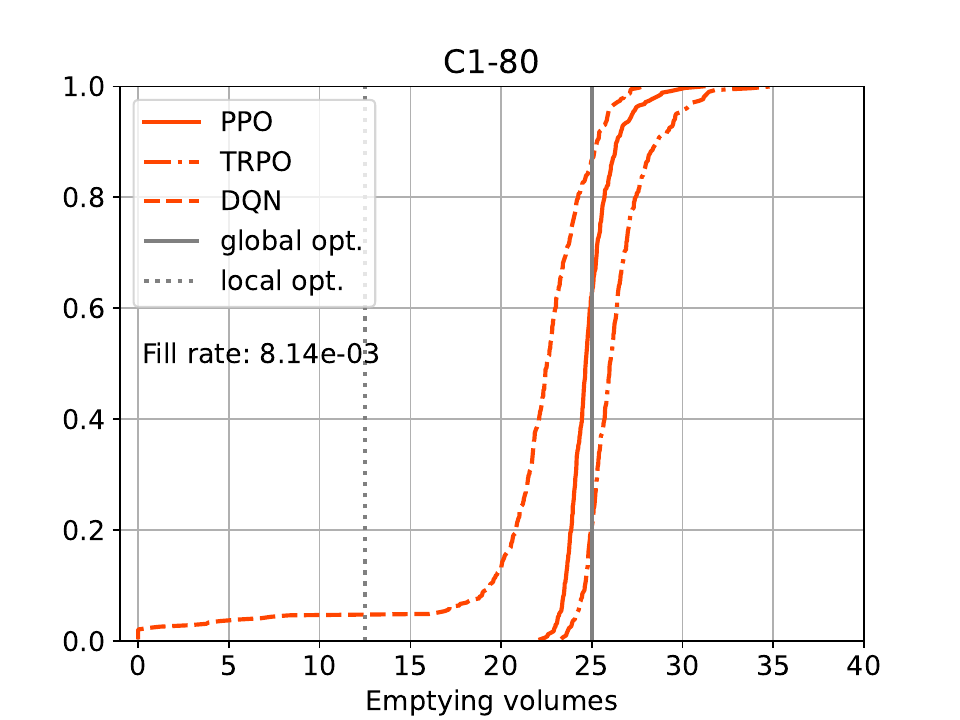} \\
\includegraphics[clip, trim=1cm 0.5cm 1cm 0.5cm, width=0.24\textwidth]{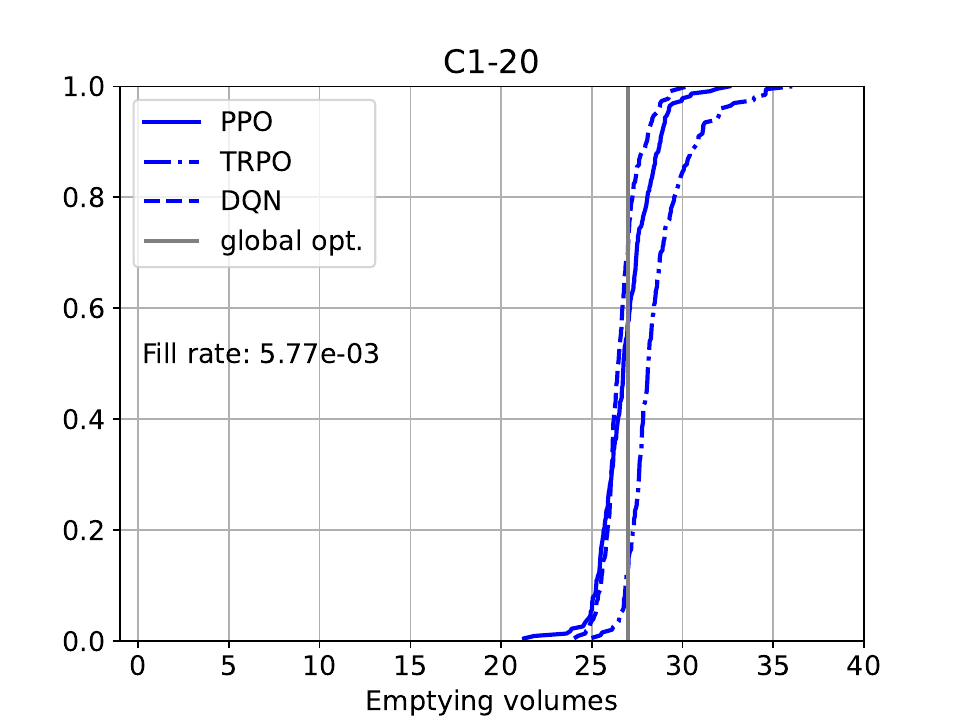}
\includegraphics[clip, trim=1cm 0.5cm 1cm 0.5cm, width=0.24\textwidth]{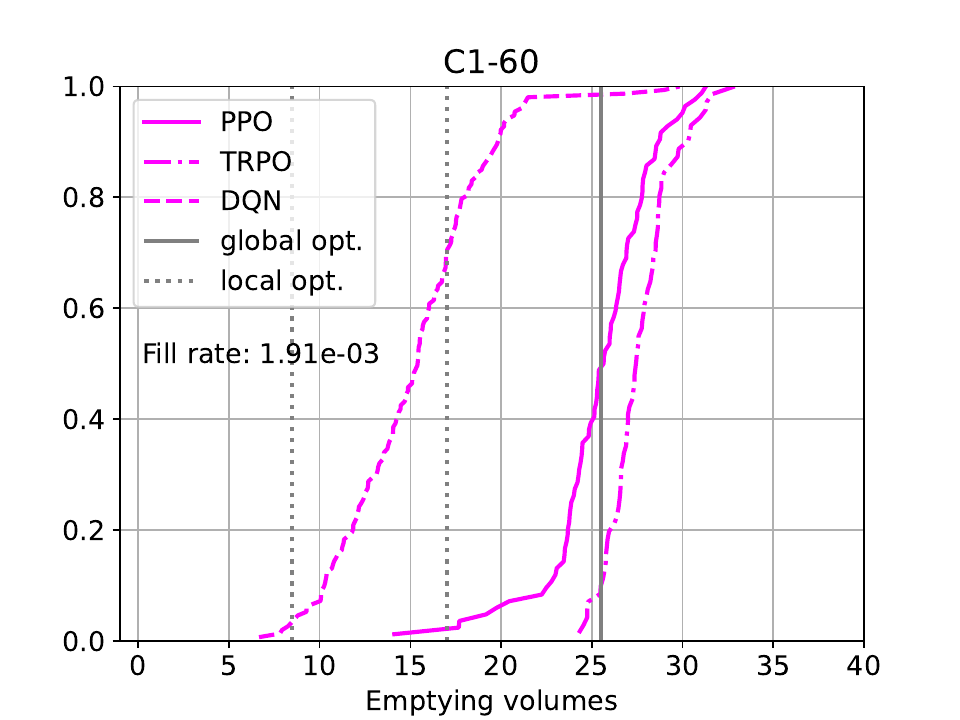}
\includegraphics[clip, trim=1cm 0.5cm 1cm 0.5cm, width=0.24\textwidth]{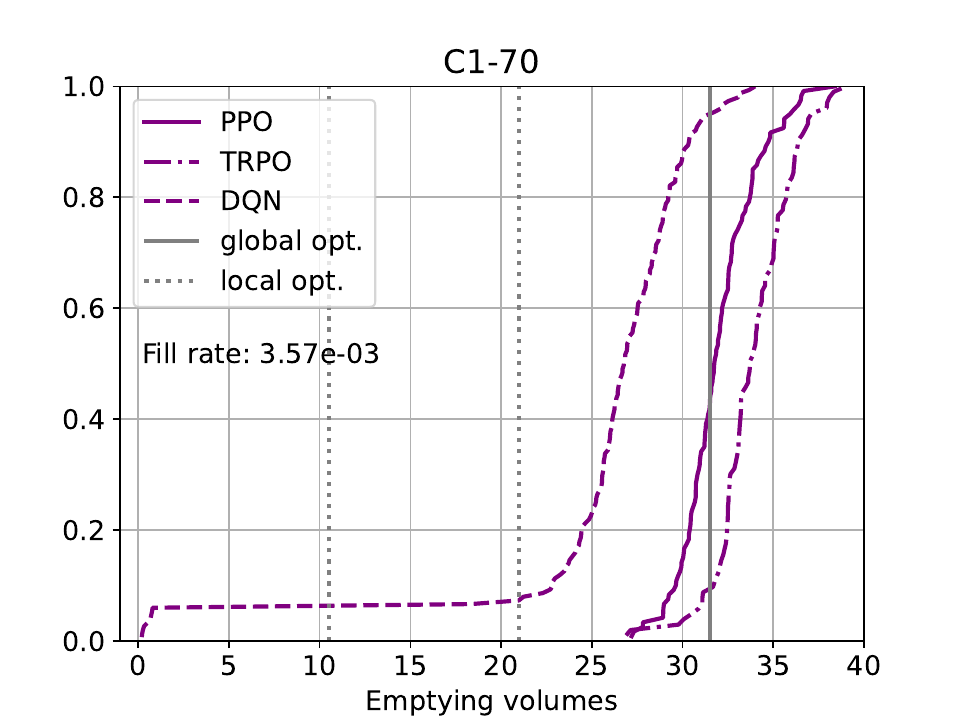}
\includegraphics[clip, trim=1cm 0.5cm 1cm 0.5cm, width=0.24\textwidth]{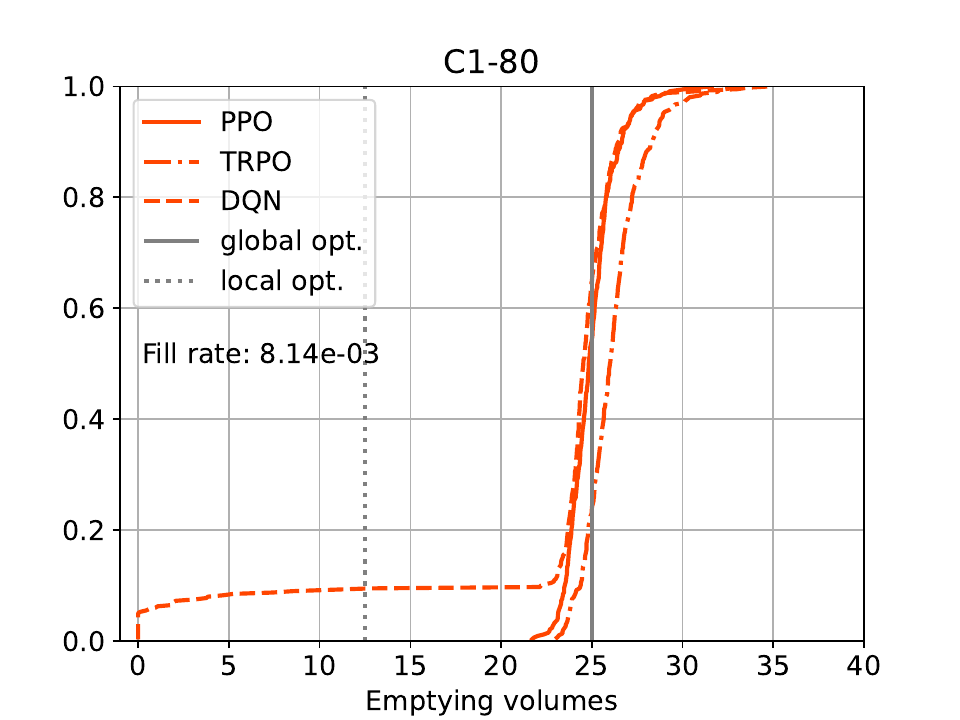}
\caption{ECDFs of emptying volumes collected over $15$ rollouts of the best policy for PPO, TRPO, and DQN on a test environment with $n = 5$ and $m = 2$. Shown are $4$ containers out of $5$. Fill rates are indicated in volume units per second. Top: training environment with $\timeincrement = 60$. Bottom: training environment with $\timeincrement = 120$.
The derivatives of the curves are the PDFs of emptying volumes. Therefore, a steep incline indicates that the corresponding volume is frequent in the corresponding density.
\label{fig:ecdfs_n5_m2}}
\end{figure}

We observe that with $\timeincrement = 60$, PPO tends to empty containers C1-60 and C1-70 prematurely, which leads to poor rewards. These two containers have the slowest fill rate. Increasing the timestep length, however, alleviates this defect. The opposite effect is observed on C1-60 with DQN, whereas no significant container-specific behavior change is observed for TRPO, as evidenced by the cumulative rewards. To further explain these results, we investigate the empirical cumulative distribution functions (ECDFs) of the emptying volumes per container. Figure~\ref{fig:ecdfs_n5_m2} reveals that more than $90\%$ of PPO's emptying volumes on C1-60 (resp. C1-70) are approximately within a $3$ (resp. $5$) volume units distance of the third best optimum when $\timeincrement = 60$. By increasing the timestep length, the emptying volumes become more centered around the global optimum. While no such clear pattern is observed with DQN on containers with slow fill rates, the emptying volumes on C1-60 move away from the global optimum when the timestep length is increased (more than $50\%$ of the volumes are within a $3$ volume units distance of the second best optimum). DQN's performance increases when $\timeincrement = 120$. This is explained by the better emptying volumes achieved on C1-20, C1-70, and C1-80.

\begin{figure}
\centering
\begin{subfigure}[t]{0.49\textwidth}
    \centering
    \includegraphics[width=\textwidth,height=3.0cm]{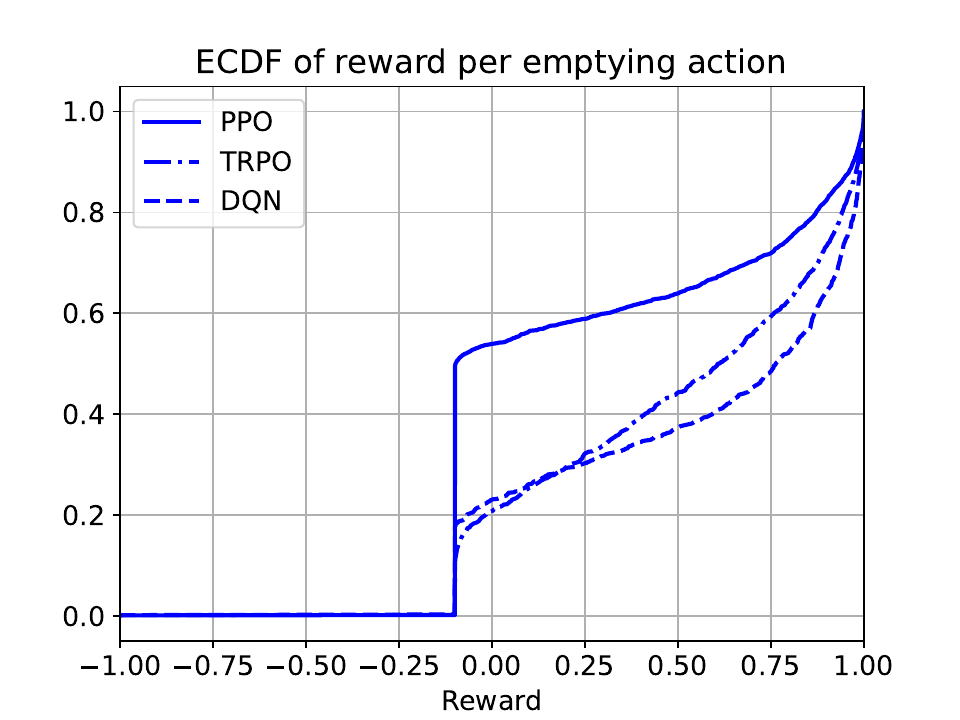}
    \caption{} \label{fig:ecdfs_reward_ppo_trpo_dqn}
\end{subfigure}
\hfill
\begin{subfigure}[t]{0.49\textwidth}
    \centering
    \includegraphics[width=\textwidth,height=3.0cm]{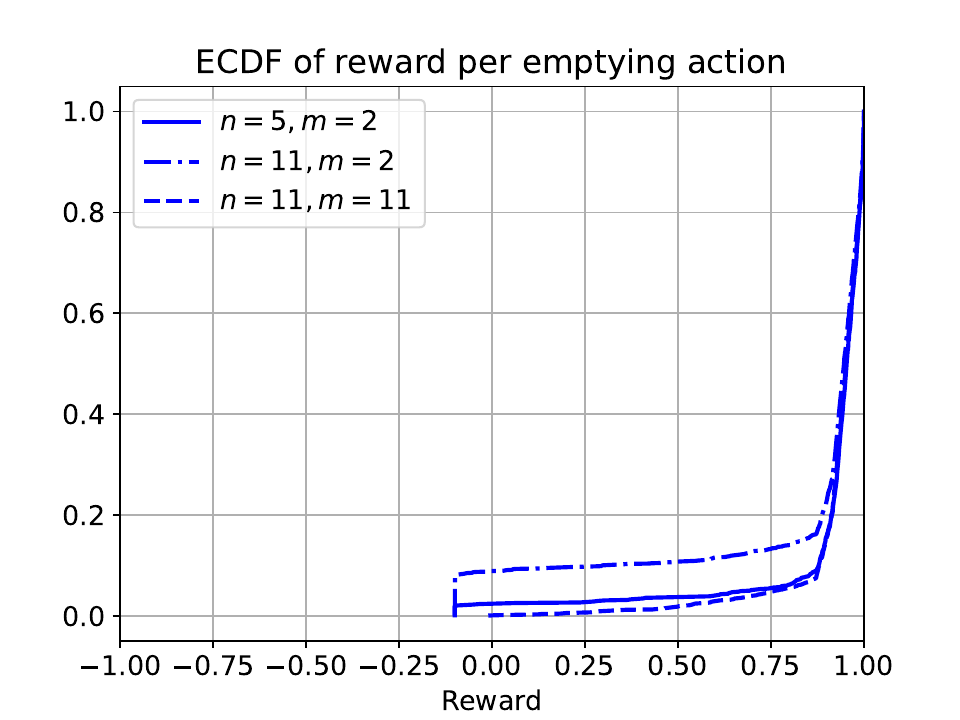}
    \caption{} \label{fig:ecdfs_reward_rule_based}
\end{subfigure}
\caption{ECDF of the reward obtained for each emptying action over $15$ rollouts. (a): Best policies for PPO, TRPO and DQN, trained on an environment with $m = n = 11$ and $\timeincrement = 120$. (b): Rule-based controller on three environment configurations.} \label{fig:ecdfs_reward}
\end{figure}

None of the benchmarked algorithms manage to learn the optimal policy when there are as many PUs as containers, independently of the environment dimensionality and the timestep length. When $m = n$, the optimal policy is known and consists in emptying each container when the corresponding global optimal volume is reached, as there is always at least one free PU. Therefore, achieving the maximum reward at each emptying action is theoretically possible. Figure~\ref{fig:ecdfs_reward_ppo_trpo_dqn} shows the ECDF of the reward per emptying action over $15$ episodes of the best policy for each of PPO, TRPO and DQN when $m = n = 11$. The rewards range from $\rmin = -1$ to $1$, and the ratio of negative rewards is particularly high for PPO. An analysis of the ECDFs of its emptying volumes (not shown) reveals that this is due to containers with slow fill rates being emptied prematurely.
TRPO achieves the least amount of negative rewards whereas all DQN's rollouts end prematurely due to a container overflowing.
Rewards close to $\rpen$ are explained by poor emptying volumes, as well as a bad allocation of the PUs ($r = \rpen$).
These findings suggest that, when $m = n$, it is not trivial for the tested RL algorithms to break down the problem into smaller, independent problems, where one container is emptied using one PU when the ideal volume is reached.

The limitations of these RL algorithms are further highlighted when compared to a naive rule-based controller that empties the first container whose volume is less than $1$ volume unit away from the ideal emptying volume.
Figure~\ref{fig:ecdfs_reward_rule_based} shows the ECDF of reward per emptying action obtained from $15$ rollouts of the rule-based controller on three environment configurations, namely $n = 5$ and $m = 2$, $n = 11$ and $m = 2$, and $n = 11$ and $m = 11$. When compared to PPO in particular ($m = n = 11$), the rule-based controller empties containers less often ($17.40\%$ of the actions vs. more than $30\%$ for PPO). Positive rewards are close to optimal (approx. $90\%$ in $[0.75, 1]$), whereas very few negative rewards are observed. These stem from emptying actions taken when no PU is available. These findings suggest that learning to wait for long periods before performing an important action may be challenging for some RL algorithms.

Critically, current baseline algorithms only learn reasonable behavior for containers operated in isolation. In the usual $m \ll n$ case, none of the policies anticipates all PUs being busy when emptying containers at their optimal volume, which they should ideally foresee and prevent proactively by emptying some of the containers earlier. Hence, there is considerable space for future improvements by learning stochastic long-term dependencies. This is apparently a difficult task for state-of-the-art RL algorithms. We anticipate that {\benchmark} can contribute to research on next-generation RL methods addressing this challenge.

\section{Conclusion}\label{sec:conclusion}
We have presented {\benchmark}, a real-world industrial RL environment. Its dynamics are quite basic, and therefore easily accessible. It is easily scalable in complexity and difficulty.

Its characteristics are quite different from many standard RL benchmark problems. At its core, it is a resource allocation problem. It features stochasticity, learning agents can get trapped in local optima, and in a good policy, the most important actions occur only rarely. Furthermore, implementing optimal behavior requires planning ahead under uncertainty.

The most important property of {\benchmark} is to be of direct industrial relevance. At the same time, it is a difficult problem with the potential to trigger novel developments in the future. While surely being less challenging than playing the games of Go or Starcraft II at a super-human level, {\benchmark} is still sufficiently hard to make state-of-the-art baseline algorithms perform poorly. We are looking forward to improvements in the future.

To fulfill the common wish of industrial stakeholders for explainable ML solutions, we provide insights into agent behavior and deviations from optimal behavior that go beyond learning curves. While accumulated reward hides the details, our environment provides insights into different types of failures and hence gives guidance for routes to further improvement.

\paragraph{Acknowledgements.}
This work was funded by the German federal ministry of economic affairs and climate action through the ``ecoKI'' grant.
\bibliographystyle{splncs04}
\bibliography{bibliography}
\end{document}